\pdfoutput=1

\documentclass[11pt]{article}

\usepackage[]{emnlp2021}

\usepackage{times}
\usepackage{tabu}
\usepackage{latexsym}
\usepackage{amssymb}
\usepackage{graphicx}
\usepackage{booktabs}
\usepackage{multirow, tabularx}
\usepackage{caption}
\usepackage{subcaption}
\usepackage{amsmath}
\usepackage[export]{adjustbox}

\usepackage[T1]{fontenc}

\usepackage[utf8]{inputenc}

\usepackage{microtype}

\title{Multiplicative Position-aware Transformer Models for Language Understanding}

\author{Zhiheng Huang \\
  AWS AI \\
  \texttt{zhiheng@amazon.com} \\\And
  Davis Liang \\
  AWS AI \\
  \texttt{liadavis@amazon.com} \\\AND
  Peng Xu \\
  AWS AI \\
  \texttt{pengx@amazon.com} \\\And
  Bing Xiang \\
  AWS AI \\
  \texttt{bxiang@amazon.com} \\  
  }

\begin{document}
\renewcommand\baselinestretch{0.94}
\baselineskip=0.93\normalbaselineskip

\maketitle
\begin{abstract}
Transformer models, which leverage architectural improvements like self-attention, perform remarkably well on Natural Language Processing (NLP) tasks. The self-attention mechanism is position agnostic. In order to capture positional ordering information, various flavors of absolute and relative position embeddings have been proposed. However, there is no systematic analysis on their contributions and a comprehensive comparison of these methods is missing in the literature. In this paper, we review major existing position embedding methods and compare their accuracy on downstream NLP tasks, using our own implementations. We also propose a novel multiplicative embedding method which leads to superior accuracy when compared to existing methods. Finally, we show that our proposed embedding method, served as a drop-in replacement of the default absolute position embedding, can improve the RoBERTa-base and RoBERTa-large models on SQuAD1.1 and SQuAD2.0 datasets.
\end{abstract}

\section{Introduction}

The introduction of pretrained models like BERT \cite{devlin2018} has led to new state-of-the-art results on various downstream tasks such as sentence classification and question answering. Variations of BERT, including RoBERTa \cite{liu2019-2}, XLNet \cite{yang2019}, ALBERT \cite{lan2019} and T5 \cite{raffel2019} have since further improved upon these results. At its core, BERT is non-recurrent and based on self-attention; in order to model the dependency between elements at different positions in the sequence, BERT relies on position embeddings. With BERT, the input embeddings are the sum of the token embeddings, segment embeddings, and position embeddings. The position embedding encodes the absolute positions from 1 to the maximum sequence length (usually 512). At each position is a learnable embedding vector. This absolute position embedding is used to model how a token at one position attends to another token at a different position. 

The RoBERTa paper \cite{liu2019} proposed the \textit{full-sentences} setup to prepare the pre-training dataset. Specifically, each input is packed with full sentences sampled contiguously from one or more documents such that the total length is at most 512 tokens. With this setup, the absolute position of a token can be arbitrary depending on chunk start positions. For a given sentence \textit{EU rejects German call to boycott British lamb} in a large corpus, the third word \textit{German} can appear in any position between 1 and 512. If this word is masked as a target word, the masked language model (MLM) task is then to predict this word conditioned on the surrounding words. In this case, the absolute positions information may not be as useful as the surrounding words' relative position distance to the target word. Motivated by this observation, the relative position embedding was proposed in \citet{shaw2018,huang2018}, in the context of encoder-decoder machine translation and music generation, respectively.  \citet{shaw2018} was subsequently modified in transformer-XL \cite{dai2019} and adopted in XLNet \cite{yang2019}. The relative position embedding in \citet{shaw2018} has been proven to be effective and was adopted in \citet{raffel2019,song2020}. More recent work on improving position embeddings can be found at \citet{huang2020,he2020,ke2020,dufter2021}.

Despite the proposal of various position embedding methods, a systematic analysis of these methods is missing in literature. Additionally, it is uncertain if simply replacing the default absolute position embedding with a relative position embedding method can improve an already strong model (e.g., RoBERTa). We attempt to answer these two questions in this paper. Our major contributions can be summarized as the following.
\begin{enumerate}
\item We implement\footnote{We will release the code soon.} existing major position embedding methods including the absolute position embedding \cite{devlin2018} and existing relative position embeddings \cite{shaw2018,raffel2019,huang2020,he2020,ke2020}. We compare these methods in term of their computational complexity and accuracy on downstream tasks including the GLUE and SQuAD datasets. 
\item Inspired by \citet{huang2020}, we propose a novel multiplicative embedding method which leads to superior accuracy when compared to existing embedding methods. 
\item We show that our novel position embedding, when compared to the default absolute position embedding, can improve RoBERTa-base and RoBERTa-large models performance on the SQuAD1.1 and SQuAD2.0 datasets. 
\end{enumerate}

\section{Related Work}

Previously, \citet{aswani2017} introduced position embeddings with dimensions matching the token embeddings (so that they can be summed). Specifically, they choose the sine and cosine functions at different frequencies:
\begin{eqnarray}
PE_{(pos, 2i)} = sin(pos/10000^{2i/d_{model}}) \\
PE_{(pos, 2i+1)} = cos(pos/10000^{2i/d_{model}})
\end{eqnarray}
where $pos$ is the position and $i$ is the embedding dimension. That is, each dimension of the position encoding corresponds to a sinusoid. The authors hypothesized that it would allow the model to easily learn to attend via relative positions, since for any fixed offset $k$, $PE_{pos+k}$ can be represented as a linear function of $PE_{pos}$. They also experimented with learned position embeddings \cite{gehring2017} and found that the two versions produced nearly identical results. BERT \cite{devlin2018} uses learnable position embeddings.

Previous work \cite{parikh2016} has introduced attention weights based on relative distance prior to BERT \cite{devlin2018}. More recently, \citet{shaw2018} demonstrated the importance of relative position representations. They presented an efficient way of incorporating relative position representations into the transformer self-attention layer. They achieved significant improvements in translation quality on two machine translation tasks. \citet{huang2018} has proposed a similar idea to incorporate the relative distance explicitly but in the music generation domain.  Transformer-XL \citet{dai2019} has modified \citet{shaw2018} to have the following two differences: 1) to introduce additional bias terms for queries; and 2) to re-introduce the use of a sinusoid formulation, in the hope that a model trained on a memory of a certain length can automatically generalize to a memory several times longer during evaluation\footnote{This was not rigorously verified in experiments.}. The proposed relative position embedding has been used in transformer-XL \cite{dai2019} and XLNet \cite{yang2019}. The relative position embedding by \citet{shaw2018} is proven to be effective and it is validated in BERT variants model training \cite{raffel2019,song2020}. 

More recently, \citet{huang2020} proposed a group of relative embedding methods ranging from simpler to more complex ones. They demonstrated that their proposed method can improve upon the BERT baseline on the SQuAD1.1 dataset. DeBERTa \citet{he2020} proposed a relative embedding method along the same lines as in \citet{shaw2018,huang2020}. They showed that the relative encoding method can improve the performance of both natural language understanding and natural language generation downstream tasks. \cite{ke2020} argued that the [CLS] tokens are virtual tokens which represent whole sentences. They proposed a new position encoding method called Transformer with Untied Position Encoding (TUPE), which has special treatment to [CLS] tokens.

In addition to the above works which are proposed in the context of transformer framework, \citet{chorowski2015} proposed a novel method of adding location-awareness to the attention mechanism in the sequence to sequence framework for automatic speech recognition (ASR). Their work is related to this paper as both attempt to integrate a location information into the self-attention mechanism.

\section{Position Embedding Methods}
In this section, we review the absolute position embedding used in the original BERT paper and major relative position embedding works including Shaw's \cite{shaw2018}, Raffel's \cite{raffel2019}, Huang's \cite{huang2020}, DeBERTa \cite{he2020} and Transformer with Untied Position Encoding (TUPE) \cite{ke2020}. 

\subsection{Self-Attention review}
\begin{figure}[!hbt]
    \centering
    \includegraphics[width=1.0\columnwidth]{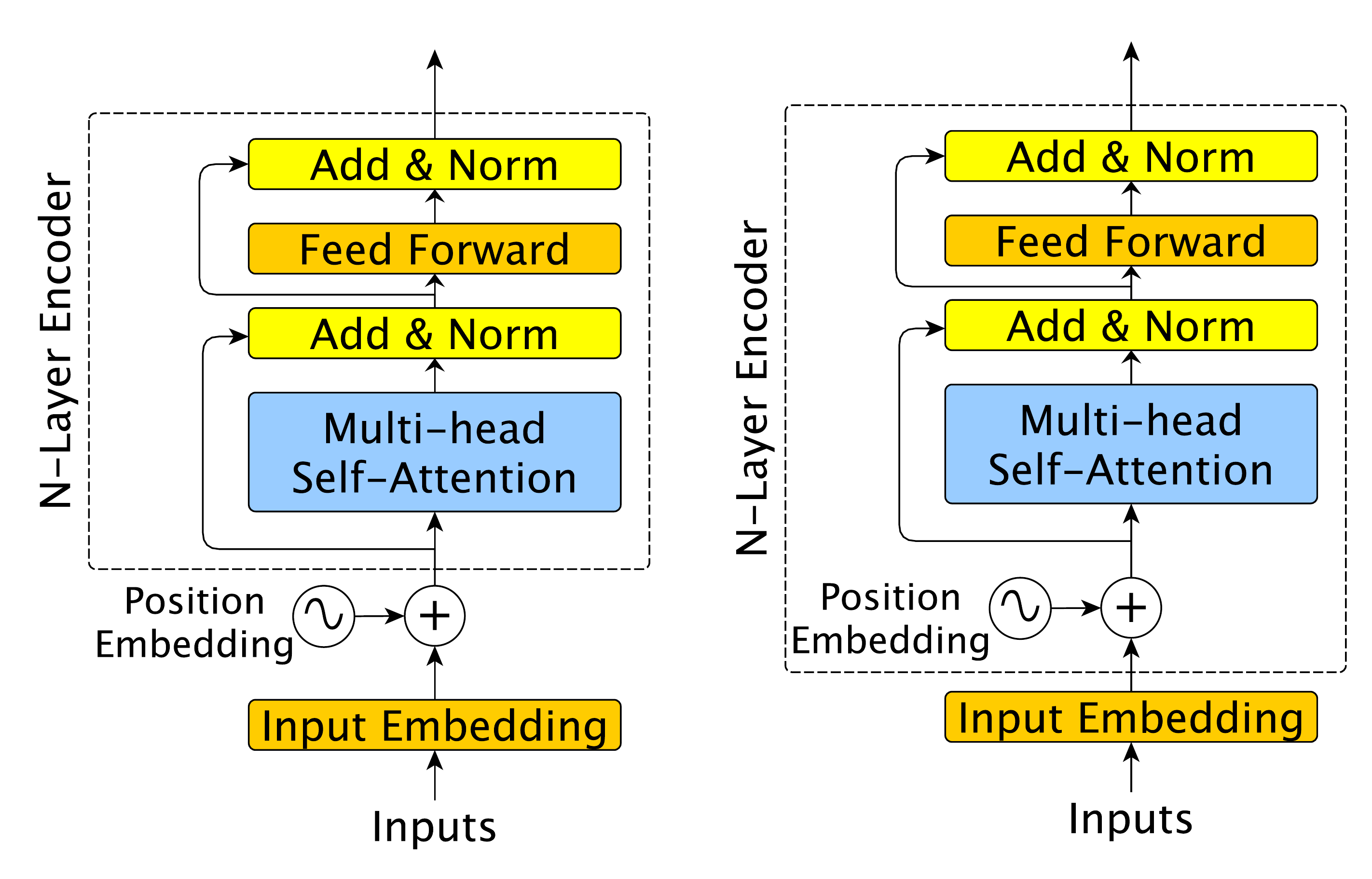}
    \caption{Transformer architectures with the original absolute position embedding (left) and all other variations of relative position embeddings (right).}
    \label{fig:trans}
\end{figure}

The BERT model consists of a transformer encoder \cite{aswani2017} as shown in Figure \ref{fig:trans}. 

The original transformer architecture uses multiple stacked self-attention layers and point-wise fully connected layers for both the encoder and decoder. Each self-attention sublayer consists of $h$ attention heads. The result from each head are concatenated to form the sublayer's output. Each attention head operates on an input sequence, $x=(x_1, \ldots, x_n)$ of $n$ elements (maximum number of tokens allowed in model training, $n$, is usually 512 by default) where $x_i \in \mathbb{R}^{d_x}$, and computes a new sequence $z = (z_1, \ldots, z_n)$ of the same length where $z_i \in \mathbb{R}^{d_z}$. Each output element, $z_i$, is computed as weighted sum of linearly transformed input elements:
\begin{equation}
z_i = \sum_{j=1}^{n} \alpha_{ij}(x_jW^V),
\end{equation}
where $\alpha_{ij}$ is the weight which is computed by applying a softmax function:
\begin{equation}
\alpha_{ij} = \frac{\exp e_{ij}}{\sum_{k=1}^n \exp e_{ik}},
\end{equation}
where $e_{ij}$ is the attention weight from position $j$ to $i$, a scaled dot product following a linear transformation:
\begin{equation}
\label{eqn:e}
e_{ij} = \frac{(x_i W^Q)(x_j W^K)^T}{\sqrt{d_z}}.
\end{equation}
The scaling factor, $\sqrt{d_z}$, is necessary to make the training stable. The dot product is chosen due to its simplicity and computational efficiency. Linear transformation of the inputs add sufficient expressive power. $W_Q$, $W_K$, $W_V \in \mathbb{R}^{d_x \times d_z}$ are parameter matrices. These parameter matrices are unique per-layer and for each attention head.

\subsection{Absolute position embedding in BERT}
\begin{figure}[!hbt]
    \centering
    \includegraphics[width=0.9\columnwidth]{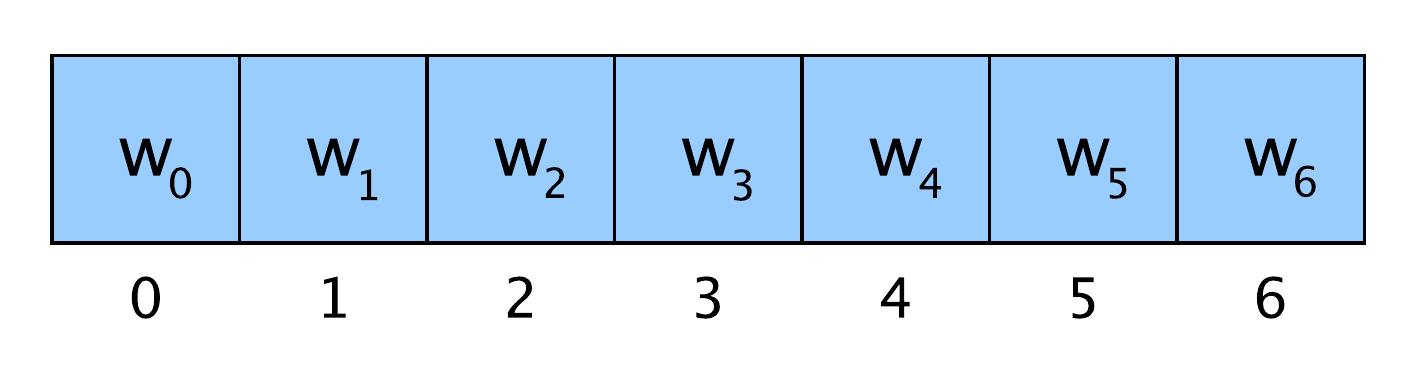}
    \caption{Absolute position embedding $p_{i}$.}
    \label{fig:absolute_weights}
\end{figure}

In the self-attention scheme, the absolute position embedding is as follows.
\begin{equation}
\label{eq:ab_pos}
x_i = t_i + s_i + w_i,
\end{equation}
where $x_i$, $i \in \{0, \ldots, n-1\}$ is the input embedding to the first transformer layer and $t_i$, $s_i$ and $w_i \in \mathbb{R}^{d_x}$ are the token embeddings, segment embeddings and absolute position embeddings, respectively.  Segment embedding indicates if a token is sentence $A$ or sentence $B$, which was originally introduced in BERT \cite{devlin2018} to compute the next sentence prediction (NSP) loss. Later work \cite{yang2019,liu2019,raffel2019} suggested that the NSP loss does not help improve accuracy. We therefore drop the segment embedding in this paper. Token embeddings $t_i$ and absolute position embeddings $w_i$, are learnable parameters trained to maximize the log-likelihood of the MLM task. Figure \ref{fig:absolute_weights} depicts the absolute position embedding graphically, which is used in the first layer in Figure \ref{fig:trans} left. The maximum length of a sequence $n$ is required to be determined before the training. Although it lacks the inductive property, this approach is found to be effective for many NLP tasks, due to the fact that the maximum sequence length is enforced at inference anyway in most cases.

\subsection{Shaw's relative position embedding}
Edge representations, $a_{ij} \in \mathbb{R}^{d_z}$, are proposed in \citet{shaw2018} to model how many token $t_i$ attends to token $t_j$. The equation (\ref{eqn:e}) can be revised as follows to consider the distance between token $i$ and token $j$ when computing their attention.
\begin{equation}
e_{ij} = \frac{(x_i W^Q)(x_j W^K + a_{ij})^T}{\sqrt{d_z}}.
\label{eqn:shaw}
\end{equation}

\begin{figure}[!hbt]
    \centering
    \includegraphics[width=0.7\columnwidth]{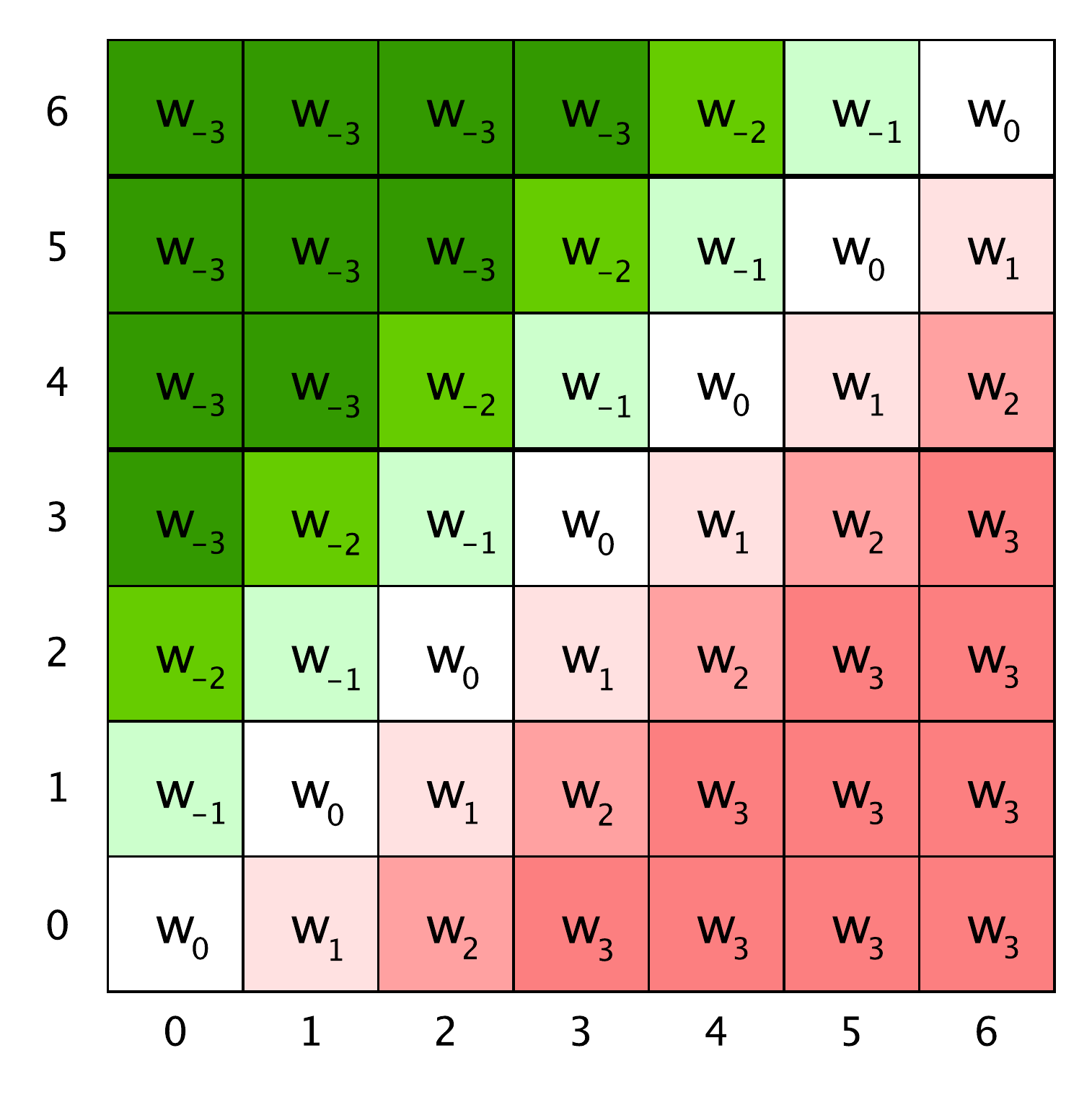}
    \caption{Relative position weights $a_{ij}$.}
    \label{fig:relative_weights}
\end{figure}

The authors also introduced a clipped value $k$ which is the maximum relative position distance allowed. The authors hypothesized that the precise relative position information is not useful beyond a certain distance. Therefore, there are $2k+1$ unique edge labels $w = (w_{-k}, \ldots, w_k)$ defined as the following.
\begin{eqnarray}
a_{ij} &=& w_{\mathrm{clip}(j-i,k)} \label{eqn:aij}\\
\mathrm{clip}(x,k) &=& \max (-k, \min(k,x))  \label{eqn:k}
\end{eqnarray}
Figure \ref{fig:relative_weights} shows the edge representations $a_{ij}$ graphically, with $k=3$. 

\subsection{Raffel's relative position embedding}
In Text-to-Text Transfer Transformer (T5), \citet{raffel2019} uses a simplified position embedding where each embedding is a scalar that is added to the corresponding logit used for computing the attention weights. Specifically, the attention weight is defined as 
\begin{equation}
e_{ij} = \frac{(x_i W^Q)(x_j W^K)^T + a_{ij}}{\sqrt{d_z}}.
\end{equation}
where $a_{ij}=w_{ij}\in\mathbb R$ are scalars used to represent how token $i$ attends to $j$. The learnable parameters are $w = (w_{1-n}, \ldots, w_{0}, \ldots, w_{n-1})$, where $n$ is the maximum sequence length. 

\subsection{Huang's relative position embedding}
Four relative embedding methods have been proposed in \cite{huang2020}. Two methods among those four, \textbf{M2} and \textbf{M4}, are selected in comparison. Similar to Raffel's method, M2 uses scalars to model relative distance and thus have an identical number of parameters. However, M2 performs multiplication instead of the addition, as in Raffel's method, when computing the attention weights.  Specifically, the attention weight defined in M2 can be written as follows. 
\begin{equation}
\label{eqn:e_2}
e_{ij} = \frac{(x_i W^Q)(x_j W^K )^T \times a_{ij}}{\sqrt{d_z}}.
\end{equation}

Huang's M4 method extends Shaw's method to include the dot product of all possible pairs of query, key, and relative position embeddings. Its attention weight is defined as follows. \begin{equation}
 \label{eqn:e_4}
 \small
e_{ij} = \frac{(x_i W^Q) \cdot (x_j W^K )  +  (x_i W^Q) \cdot a_{ij} + (x_j W^K ) \cdot a_{ij} }{\sqrt{d_z}}.
\end{equation}
Three factors in the numerator model the interaction of query and key, query and relative position embedding, and key and relative position embedding, respectively. The interaction of query and key is the conventional content attention, while the remaining two are for relative position discount of the query and key respectively. Shaw's method (equation \ref{eqn:shaw}) only contains the first two factors, which may not have as much expressive power as equation \ref{eqn:shaw}. The learnable parameters are shared in the second and third factors in the numerator, leading to the same number of parameter as Shaw's method.  

\subsection{DeBERTa relative position embedding}
Recently, DeBERTa \cite{he2020} claims to improve upon existing BERT variants. One major innovation is the disentangled attention mechanism, where each word is represented using two vectors that encode its content and position, respectively, and the attention weights among words are computed using disentangled matrices on their contents and relative positions, respectively. The attention weight, with the notations used in this paper, is listed as follows.
 \begin{equation}
 \label{eqn:deberta}
 \scriptsize
e_{ij} = \frac{(x_i W^Q) \cdot (x_j W^K )  +  (x_i W^Q) \cdot (a_{ij} W^R) + (x_j W^K ) \cdot (a_{ij} W^T) }{\sqrt{3d_z}},
\end{equation}
where $W^R, W^T \in \mathbb{R}^{d_x \times d_z}$ are additional linear projection matrices. DeBERTa's embedding method is similar to Huang's M4, with the following differences: 1) DeBERTa introduced projection matrices $W^R, W^T $ which are applied to the relative position embeddings. 2) unlike Huang's M4, in which the relative embeddings are shared, DeBERTa has different relative embeddings across different heads,  and 3) DeBERTa used a different scaling factor $\sqrt{3d_z}$, instead of the default $\sqrt{d_z}$, in the denominator.

\subsection{Transformer with untied positional encoding (TUPE)}
The self-attention mechanism employed by TUPE \cite{ke2020} computes the word contextual correlation and positional correlation separately with different parameterizations before adding them together. 
The attention weight definition, which consists of both absolute and relative embeddings, is listed as follows. 
 \begin{equation}
 \label{eqn:rethink}
 \small
e_{ij} = \frac{(x_i W^Q) \cdot (x_j W^K )  +  (p_i U^Q) \cdot (p_j U^K) }{\sqrt{2d_z}} + a_{ij},
\end{equation}
where $p_i$ and $p_j$ are the absolute position embeddings, and $U^Q, U^K \in \mathbb{R}^{d_z \times d_z}$ are the linear projection matrices for absolute position embeddings. $a_{ij}$ is used to model relative position embeddings, which is identical to Raffel's relative embeddings.

Furthermore, the authors argued that the [CLS] token is a virtual token and it should not be applied in the same way to the normal tokens. Otherwise, [CLS] is likely biased to focus on the first several words instead of the whole sentence, hurting the performance on downstream tasks. Motivated by this, TUPE \cite{ke2020} proposed the following attention weight.
 \begin{equation}
 \label{eqn:rethink}
 \small
 \begin{split}
e_{ij} & =  \frac{(x_i W^Q) \cdot (x_j W^K )}{\sqrt{2d_z}}  \\
         & +   \text{reset}_\theta(\frac{(p_i U^Q) \cdot (p_j U^K)  }{\sqrt{2d_z}} + a_{ij}, i, j),
 \end{split}
\end{equation}
where $\text{reset}_\theta$ is defined as the following, in which $\theta_1$ is used as the attention weight from [CLS] token to others, and $\theta_2$ is used from others to [CLS]. In other words, [CLS] tokens use special relative position embeddings instead of the embeddings assigned by relative distances.
\begin{eqnarray}
 \label{eq:reset}
 \text{reset}_\theta(v_{ij}, i, j) = \left\{ \begin{array}{ll}
v_{ij} & i \ne 1, j \ne 1\\
\theta_1 & i = 1\\
\theta_2 & i \ne 1, j = 1
\end{array} \right.
\end{eqnarray}

\subsection{Proposed M4 multiplicative (M4M) method}
The M2 method proposed in \cite{huang2020} essentially replaces the additive operator with multiplicative operator in Raffel's method, which improves downstream application accuracy as we will see in the experiment section \ref{sec:evaluation}. It is intuitive to apply the multiplicative operator to scale the content attention with respect to the relative positional attention. Inspired by this intuition, we propose to replace the additive operator with multiplicative operator in M4 method (equation \ref{eqn:e_4}) to the following. 
 \begin{equation}
 \label{eqn:e_5}
 \small
e_{ij} = \frac{(x_i W^Q) \cdot (x_j W^K )  \times  (x_i W^Q) \cdot a_{ij}  \times (x_j W^K ) \cdot a_{ij} }{\sqrt{d_z}}.
\end{equation}
We denote this method as M4M, which has exactly the same amount of parameters as M4, leading to an identical computational complexity.

\section{Complexity Analysis}
We analyze the storage complexity of various position embedding methods in this section. For a transformer model with $m$ layers, $h$ attention heads per layer, and maximum sequence length of $n$, table \ref{tab:complexity} lists the parameter size for various position embeddings and the runtime storage complexity.  In order to have sufficient expressive power, we allow different embedding parameters at different layers for all methods  (see Figure \ref{fig:trans} right) except absolute position embedding\footnote{To be compatible to the original BERT implementation.}. For example, Shaw's method introduces the following parameters size:  $m(2n-1)d/h=12*(2*512-1)*768/12=785K$. It has runtime storage complexity of $\mathcal{O}(mn^2d/h)$. 
\begin{table*}[!hbpt]
    \centering
    \begin{tabular}{lccc}
    Method & Parameters & Parameter (K) & Complexity \\ \hline
    Absolute & $nd$ & 393 & $\mathcal{O}(nd)$ \\
    Shaw & $m(2n-1)d/h$ & 785 & $\mathcal{O}(mn^2d/h)$ \\
    Raffel & $m(2n-1)$ & 12 & $\mathcal{O}(mn^2)$ \\
    M2 & $m(2n-1)$ & 12 & $\mathcal{O}(mn^2)$ \\
    M4 & $m(2n-1)d/h$ & 785 & $\mathcal{O}(mn^2d/h)$ \\
    DeBERTa & $m(2n-1)d/h + m(d/h)^2$ & 834 & $\mathcal{O}(mn^2d/h)$ \\
    TUPE & $mnd/h + m(d/h)^2 + m(2n-1)$ & 454 & $\mathcal{O}(mn^2d/h)$ \\
    M4M & $m(2n-1)d/h$ & 785 & $\mathcal{O}(mn^2d/h)$ \\
    \hline
    \end{tabular}
    \caption{Parameter sizes and runtime storage complexities of various position embedding methods. }
\label{tab:complexity}
\end{table*}
All position embedding methods introduce a small number of additional parameters to the BERT model. The maximum parameter size is 834K from DeBERTa. It is negligible when compared to the number of parameters in RoBERTa-base (110M parameters). Other methods introduce even fewer parameters. For example, Raffel and M2 only have 12K additional parameters. In terms of training and inference speed, all methods are similar to the absolute position embedding baseline. Note that since TUPE requires different treatment to [CLS] positions, the self-attention along batch dimension cannot be efficiently computed with broadcasting, thus slowing down the training speed. 

\section{Experiments}

\subsection{Experimental setup} 

We start with a small-scale pretraining setup so we can compare different embedding methods effectively. We use the training data of BooksCorpus \cite{zhu2015} and English Wikipedia \cite{wiki2004,devlin2018} (15G in total), which is the dataset used to pretrain original BERT model \cite{devlin2018}.  Following the setup from RoBERTa \cite{liu2019}, we leave out the next sentence prediction loss and only use one segment instead of two segments as proposed in BERT during model training. We set the maximum input length to 512. We use the RoBERTa-base setting: 12 layers, each layer has 768 hidden units and 3072 intermediate units. We use the same vocabulary as used in RoBERTa. The model updates use a batch size of $96$ and Adam optimizer with learning rate starting at 1e-4. This batch size was chosen such as to fit as many samples as possible, constrained by GPU memory. We train each model on one AWS P4DN instance (each containing 8 A100-SXM4-40GB GPUs) with a maximum steps of 500000, which corresponds to 5.85 epochs.  Each pretraining run takes approximately 30 hours. Following previous work \citet{devlin2018,yang2019,liu2019,lan2019}, we evaluate on the General Language Understanding Evaluation (GLUE) benchmark \cite{wang2018} and the Stanford Question Answering Dataset (SQuAD1.1 and SQuAD2.0) \cite{rajpurkar2016}. 

\subsection{Unsuccessful baselines} 
We first train a RoBERTa-base model without any position embedding information. Specifically, the input is the sum of the token id embedding and token type id embedding with no absolute position embedding. We find that the model pre-training does not converge. The training loss ended up being around 5.8\footnote{Normally a RoBERTa-base model reaches the training loss of 1.4.} which does not result in good accuracy on downstream applications. 

We use the full-sentences setting in \citet{liu2019}. Each input is packed with full sentences sampled contiguously from one or more documents, such that the total length is at most 512 tokens. With the absolute position embedding, each token has a position in the range of 1 to 512. Due to sentence packing, the beginnings of sentences can be associated with any positions. While this absolute position embedding leads to a reasonable baseline as we will see in Section \ref{sec:evaluation}, we train a RoBERTa model with the \textit{real} absolute positions. That is, each token is associated with its real position in the sentence which it belongs to. With this setup, the first token of a sentence always has the position of 1. We hypothesize that the real absolute position may be useful to capture certain syntactic information in model training. Surprisingly, we find that the model pre-training does not converge; the training loss remains as around 4.3 after the pre-training.

\subsection{Evaluation on different embedding methods} 
\label{sec:evaluation}
We train RoBERTa-base models with our implementation of different position embedding methods including absolute, Shaw, Raffel, M2, M4, DeBERTa, TUPE and M4M. Table \ref{tab:compare} shows the accuracy of different embedding methods on MNLI, SST-2 and SQuAD 1.1 dataset. Absolute is the default absolute position embedding which results in the accuracy of 81.25 and 90.25 on MNLI and SST-2 datasets, respectively. It produces an F1 score of 87.08 on the SQuAD1.1 dataset. The lower F1 score when compared to the BERT \cite{devlin2018} baseline of 88.5 can be attributed to under-training. While training a BERT model requires around 40 epochs, our pre-training experiment consists of approximately 6 epochs. Shaw's method offers accuracy boost over the absolute embedding on all three datasets. For example, it improves performance on the SQuAD1.1 dev, increasing the F1 from 87.08 to 88.62. M2 uses fewer parameters (12K) compared to Shaw's method (785K) and it results in a lower accuracy. M4 has the same number of parameters as Shaw's method. It outperforms Shaw's method on MNLI and SQuAD1.1 but under performs on the SST-2 dataset. DeBERTa has mixed result when compared to M4. It improves over the M4 method on MNLI dataset but lowers accuracy on both the SST-2 and SQuAD1.1 datasets. The proposed M4M method leads to accuracy improvement for all three datasets when compared to M4. Specifically, it achieves the highest accuracy (91.62) on the SST-2 dataset and F1 score (89.45) on the SQuAD1.1 among all methods. 
\begin{table}[!hbpt]
    \centering
    \small
    \begin{tabular}{lccc}
    \toprule
    \textbf{Model} & \textbf{MNLI-m} & \textbf{SST-2} & \textbf{SQuAD1.1}   \\ 
    \midrule
    Absolute &  81.25  & 90.25    & 87.08   \\
    Shaw  & 82.88  &  91.28   &  88.62 \\
    M2 & 82.41  &  90.36  &  87.26 \\
    M4& 83.05   &  91.05  &  89.36 \\
    DeBERTa & \textbf{83.67}  & 90.71   &  88.84 \\
    M4M&  83.58 & \textbf{91.62}   & \textbf{89.45}  \\
    \midrule
    M4+Reset &  82.30 &  91.39  &  88.43 \\
    ABS+M4M& 83.19  &  91.16  &  88.68 \\
    \bottomrule
    \end{tabular}
    \caption{Accuracy of different position embeddings for RoBERTa-base. }
\label{tab:compare}
\end{table}

We also implemented Raffel's method which has exactly the same amount of parameters as the M2 method. We observe that the training loss of Raffel is similar to other methods. However, the fine-tuning diverges on GLUE and SQuAD1.1 datasets if the same learning rate is used. The fine-tuning is able to converge with a smaller learning rate, for example, 2.5e-5 on SQuAD1.1 (instead of the default 3e-5). However, the final accuracy on SQuAD1.1 is 83.43 which is worse than others. TUPE \cite{ke2020} also uses Raffel's relative position embedding and it suffers from the same issue. We thus do not include Raffel and TUPE results in Table \ref{tab:compare}.

Finally, we implemented two hybrid methods: M4+Reset and ABS+M4M. M4+Reset applies the reset equation (\ref{eq:reset}) to M4 method. That is, the [CLS] token has its own relative embedding in the M4 method. We implement this to test if the reset mechanism in \citet{ke2020} can help boost accuracy on downstream applications. However, we cannot validate this hypothesis as M4+Reset overall under performs M4 on all three datasets. In addition, due to the special treatment of [CLS] tokens, the self-attention along batch dimension cannot be efficiently computed with broadcasting, thus slowing down the training speed. We also implemented ABS+M4M which includes both absolute and relative position information. This combination under performs M4M method on all three test datasets, which suggests that the combination introduces more noise than useful complementary information.

\subsection{The effect of scaling factor}
In original transformer models, the scaling factor of $\sqrt{d_z}$  was used in equation (\ref{eqn:e}) to make training stable. Different scaling factors have been introduced in the self-attention equations, for example, $\sqrt{3d_z}$ is used in equation (\ref{eqn:deberta}) for DeBERTa and $\sqrt{2d_z}$ is used in equation (\ref{eqn:rethink}) for TUPE. The effect of scaling has not been evaluated in any previous study.  In this section, We vary the scaling factors to the M4 methods and report the downstream task accuracy in Table \ref{tab:scaling_factor}.  The results show that different scaling factors do not make a significant accuracy difference on the MNLI, SST-2 and SQuAD1.1 datasets. 

\begin{table}[!hbpt]
    \centering
    \small
    \begin{tabular}{lccc}
    \toprule
    \textbf{Scaling Factor} & \textbf{MNLI-m} & \textbf{SST-2} & \textbf{SQuAD1.1}   \\ 
    \midrule
    1  & 83.05   &  91.05  &  \textbf{89.36} \\
    2  &  82.80 & \textbf{91.97}   &  88.96  \\
    3 &  \textbf{82.86}  &  91.51  &  89.20 \\
    4  & 82.58  &  91.28 &  \textbf{89.36}  \\
    6  &  82.71  &  90.71  &  88.95 \\
    9  & 82.34  &  90.36  &  88.65  \\
     \bottomrule
    \end{tabular}
    \caption{Accuracy of M4 method with different scaling factors. }
\label{tab:scaling_factor}
\end{table}

\subsection{The effect of sharing parameters}
In our implementation of various relative position embedding methods, we share the parameters across different heads for each layer. That is why we see the term of $d/h$ instead of $d$ in the \textit{parameters} column in Table \ref{tab:complexity} for Shaw, M4, DeBERTa, TUPE and M4M methods. Some methods, for example DeBERTa \cite{he2020}, allow different relative embeddings across different heads. In this section, we train two RoBERTa base models with the M4 method, one with parameter sharing and one without.  Table \ref{tab:parameter_sharing} shows their accuracy on the MNLI, SST-2 and SQuAD1.1 datasets. As we can see, the parameter sharing performs better on two out three datasets. 
\begin{table}[!hbpt]
    \centering
    \small
    \begin{tabular}{lccc}
    \toprule
    \textbf{Parameter Sharing} & \textbf{MNLI-m} & \textbf{SST-2} & \textbf{SQuAD1.1}   \\ 
    \midrule
    Yes  & \textbf{83.05}   &  91.05  &  \textbf{89.36} \\
    No  &  82.88 & \textbf{91.17}   &  88.50  \\
    \bottomrule
    \end{tabular}
    \caption{Accuracy of M4 method with and without parameter sharing. }
\label{tab:parameter_sharing}
\end{table}

\subsection{RoBERTa-base model with M4M method} \label{RoBERTa-base-compare}
\begin{table*}[!hbpt]
    \centering
    \small
    \begin{tabular}{lcccccccc}
    \toprule
     Model & MNLI-(m/mm) & QQP & QNLI & SST-2 &CoLA& STS-B& MRPC &  RTE   \\ 
     & 392k & 363k & 108k & 67k & 8.5k & 5.7k & 3.5k & 2.5k \\
    \midrule
    RoBERTa &  87.32/87.31  &  \textbf{88.33}   & 92.84 & 93.92   & \textbf{55.33} & \textbf{90.33} & 91.42  & 71.84 \\    
    RoBERTa-ABS & 87.35/86.99 & 88.29  & 92.78  & 93.92 & 52.77 & 89.07  & 89.67 & \textbf{72.92} \\
    RoBERTa-M4M  & \textbf{87.82}/\textbf{87.59}  & 88.28  & \textbf{92.98} &  \textbf{94.26} & 47.02   & 89.45 &  \textbf{91.63} & 68.59 \\  
     \bottomrule
    \end{tabular}
    \caption{GLUE accuracy for RoBERTa, RoBERTa-ABS, and RoBERTa-M4M models. The number below each task denotes the number of training examples. F1 scores are reported for QQP and MRPC, Spearman correlations are reported for STS-B, and accuracy scores are reported for the other tasks. }
\label{tab:glue}
\end{table*}
We compared different embedding methods in previous sections. We now apply the best method, M4M, to see if it can lead to an accuracy boost using a larger scale training setup. We now use the training data of BooksCorpus \cite{zhu2015} plus English Wikipedia \cite{wiki2004,devlin2018} (16G) and OpenWeb Text \cite{gokaslan2019} (38G). The CommonCrawl News dataset \cite{nagel2016} (76G) and STORIES \cite{trinh2018} (31G) datasets, which were used in RoBERTa pre-train, are not included in our pre-training due to unavailability. As we do not have exactly the same training dataset as RoBERTa, we compare three models in our experiments, \textit{RoBERTa}, \textit{RoBERTa-ABS}, and \textit{RoBERTa-M4M} to test the effectiveness of the proposed M4M method. RoBERTa is the official RoBERTa model\footnote{\textit{roberta-base} from pytorch transformer https://github.com/huggingface/transformers.}. RoBERTa-ABS and RoBERTa-M4M are the models pre-trained with absolute position embedding and M4M respectively. Both are initialized from RoBERTa and are trained with the following setup. We use 4 AWS P4DN instances (each has 8 A100-SXM4-40GB GPUs) for each model pre-training. We use a batch size of $480$ and Adam optimizer with learning rate starting from 1e-4. We train each model with maximum steps of 500000 (approximately 8.8 epochs). The pre-training of RoBERTa-ABS or RoBERTa-M4M takes around 5 days to complete.

Following \citet{devlin2018}, we use a batch size of 32 and 3-epochs of fine-tuning for each dataset in GLUE. For each task, we report the best accuracy on the development dataset with learning rates  2e-5, 3e-5 and 4e-5. Table \ref{tab:glue} shows the results of GLUE datasets for RoBERTa, RoBERT-ABS, and RoBERTa-M4M. Due to the fact that we use less data in pretraining (54G vs 161G), RoBERT-ABS underperforms the official RoBERTa model on 6 out 8 datasets. Using the same, smaller pretraining corpus (16G) as we did with RoBERTa-ABS, RoBERTa-M4M is able to match the performance of the official RoBERTa model, mainly due to the effectiveness of the M4M method. It outperforms RoBERTa on the MNLI, QNLI, SST-2 and MRPC datasets but underperforms on the rest of the four datasets. It is worth noting that RoBERTa-M4M tend to outperform RoBERTa when the training dataset is relatively large. We hypothesize that, with more pretraining and downstream training data, the RoBERTa-M4M may obtain further improvements in accuracy.

We also test RoBERTa, RoBERTa-ABS, and RoBERTa-M4M on the SQuAD1.1 and SQuAD2.0 datasets. Table \ref{tab:squad} shows that RoBERTa-ABS boosts RoBERTa performance slightly. For example, it improves F1 score from 83.10 to 83.29 on SQuAD2.0 datasets. RoBERTa-M4M results in an even greater accuracy boost. It leads to the highest accuracies among three models on both the SQuAD1.1 and SQuAD2.0 datasets. For example, it achieves F1 score of 84.00 on the SQuAD2.0 dataset. 

Finally, we train a RoBERTa-M4M (denoted as RoBERTa-M4M (C4 en) in Table \ref{tab:squad}) on C4-en dataset \footnote{https://huggingface.co/datasets/allenai/c4/tree/main.}, which has 800G English text and is much larger than the combination of BooksCorpus, English Wikipedia, and OpenWebText (54G). RoBERTa-M4M (C4 en) leads to additional accuracy gain when compared to RoBERTa-M4M. It achieves the best accuracy on both SQuAD1.1 and SQuAD2.0 datasets. For example, it achieves F1 score of 84.51 on SQuAD2.0 dataset.
\begin{table}[!hbpt]
    \centering
    \small
    \begin{tabular}{lcccc}
    \toprule
    Model & \multicolumn{2}{c}{\textbf{SQuAD1.1}} & \multicolumn{2}{c}{\textbf{SQuAD2.0} } \\
     & EM & F1
    & EM & F1  \\
    \midrule
    RoBERTa  & 85.95 &  92.12   & 79.99 & 83.10 \\
    RoBERTa-ABS & 86.10 & 92.31 & 80.67 & 83.69\\
    RoBERTa-M4M &   \textbf{86.44} & \textbf{92.52}  & \textbf{80.88} & \textbf{84.00} \\
    \midrule
    RoBERTa-M4M (C4 en) &   \textbf{86.54} & \textbf{92.64}  & \textbf{81.65} & \textbf{84.51} \\
        \bottomrule
    \end{tabular}
    \caption{SQuAD1.1 and SQuAD2.0 accuracy for RoBERTa, RoBERTa-ABS, RoBERTa-M4M, RoBERTa-M4M (C4 en) models. }
\label{tab:squad}
\end{table}

\subsection{RoBERTa-large model with M4M method}
Similar to the base model comparison in Section \ref{RoBERTa-base-compare}, we assess how the proposed M4M method affect RoBERTa \textit{large} models in this section. We use the training data of BooksCorpus \cite{zhu2015} plus English Wikipedia \cite{wiki2004,devlin2018} (16G) and OpenWeb Text \cite{gokaslan2019} (38G) to train a model denoted as \textit{RoBERTa-M4M Large}, which is initialized from RoBERTa Large \footnote{\textit{roberta-large} from pytorch transformer https://github.com/huggingface/transformers.}. We use 4 AWS P4DN instances (each has 8 A100-SXM4-40GB GPUs) for model pre-training. We use the batch size of $192$, Adam optimizer with learning rate starting from 1e-4, and maximum steps of 500000. 

We test RoBERTa Large and RoBERTa-M4M Large on the SQuAD1.1 and SQuAD2.0 datasets. Table \ref{tab:squad_large} shows that RoBERTa-M4M Large can improve RoBERTa Large model. Specifically, it improves from F1 score of 94.63 to 94.78 on SQuAD1.1 dataset, and from F1 score of 87.62 to 88.34 on SQuAD2.0 dataset. 
\begin{table}[!hbpt]
    \centering
    \small
    \begin{tabular}{lcccc}
    \toprule
    Model & \multicolumn{2}{c}{\textbf{SQuAD1.1}} & \multicolumn{2}{c}{\textbf{SQuAD2.0} } \\
     & EM & F1
    & EM & F1  \\
    \midrule
    RoBERTa Large  & 89.13 &  94.63   & 84.66 & 87.62 \\
    RoBERTa-M4M Large &   \textbf{89.21} & \textbf{94.78}  & \textbf{85.47} & \textbf{88.34} \\
    \bottomrule
    \end{tabular}
    \caption{SQuAD1.1 and SQuAD2.0 accuracy for RoBERTa Large and RoBERTa-M4M Large models. }
\label{tab:squad_large}
\end{table}

\section{Conclusion}
We reviewed major existing position embedding methods and compared their accuracy on downstream NLP tasks using our own implementations. We also proposed a novel multiplicative embedding method which leads to superior accuracy when compared to existing methods. Finally, we showed that our proposed embedding method can improve both RoBERTa-base and RoBERTa-large models on SQuAD1.1 and SQuAD2.0 datasets.

\bibliographystyle{acl_natbib}
\bibliography{emnlp2021}




\end{document}